\documentclass{zgroup/zgroup}

\usepackage{amsmath,amsfonts,bm}

\def\eqref#1{equation~\ref{#1}}

\def\1{\bm{1}}

\DeclareMathAlphabet{\mathsfit}{\encodingdefault}{\sfdefault}{m}{sl}
\SetMathAlphabet{\mathsfit}{bold}{\encodingdefault}{\sfdefault}{bx}{n}

\usepackage[utf8]{inputenc}
\usepackage{booktabs}
\usepackage{hyperref}
\usepackage{graphicx, color}

\usepackage{times}
\usepackage{url}
\usepackage{microtype}
\usepackage{appendix}
\usepackage{caption}

\usepackage{amsfonts}       
\usepackage{nicefrac}       
\usepackage{microtype}      
\usepackage{multirow, multicol}
\usepackage{ragged2e}
\usepackage{amsmath}
\usepackage{thmtools, thm-restate}
\usepackage{bm, bbm}
\usepackage{enumitem}
\usepackage{tikz}
\usepackage{wrapfig}
\usetikzlibrary{positioning}
\usepackage{lipsum}
\usepackage{float}
\usepackage{longtable}

\title[Test-Time Warmup for MLLMs]{Test-Time Warmup for Multimodal Large Language Models}

\author
{\Name{Nikita Rajaneesh}
\Email{nr2937@columbia.edu}\\
\addr Columbia University\\
\Name{Thomas Zollo}
\Email{tpz2105@columbia.edu}\\
\addr Columbia University \\
\Name{Richard Zemel}
\Email{zemel@cs.columbia.edu}\\
\addr Columbia University \\}
\date{October 2020}

\begin{document}

\maketitle

\begin{abstract}
    Multimodal Large Language Models (MLLMs) hold great promise for advanced reasoning at the intersection of text and images, yet they have not fully realized this potential. MLLMs typically integrate an LLM, a vision encoder, and a connector that maps the vision encoder’s embeddings into the LLM’s text embedding space. Although each component is pretrained on massive datasets with billions of samples, the entire multimodal model is typically trained on only thousands (or a few million) samples, which can result in weak performance on complex reasoning tasks. To address these shortcomings, instead of relying on extensive labeled datasets for fine-tuning, we propose a Test-Time Warmup method that adapts the MLLM per test instance by leveraging data from weakly supervised auxiliary tasks. With our approach, we observe a relative performance improvement of 4.03\% on MMMU, 5.28\% on VQA-Rad, and 1.63\% on GQA on the Llama-Vision-Instruct model. Our method demonstrates that ‘warming up’ before inference can enhance MLLMs’ robustness across diverse reasoning tasks.
\end{abstract}

\section{Introduction}

In recent years, large language models (LLMs) have taken center stage in natural language processing, achieving remarkable results across a wide range of tasks \citep{minaee2025largelanguagemodelssurvey, brown_lms_few_shot_learners, rae2022scalinglanguagemodelsmethods,touvron2023llama2openfoundation}. Concurrently, vision foundation models such as CLIP have redefined image understanding \citep{radford2021learningtransferablevisualmodels}. Building on these advances, multimodal large language models (MLLMs) (\citep{liu2023visualinstructiontuning, grattafiori2024llama3herdmodels}) have shown promising performance in multimodal tasks such as image captioning \citep{li2023blip2bootstrappinglanguageimagepretraining} and visual question answering \citep{mckinzie2024mm1methodsanalysis}, offering a vision of a future where advanced reasoning is robust across text and image. 

MLLMs typically integrate an LLM, a vision encoder (e.g., CLIP), and a connector that maps the vision encoder's embeddings into the LLM text embedding space. Despite the power of their individual components, each trained on billions of samples, for the final instruction tuning step, MLLMs are often limited by a relatively small amount of multimodal training data, typically ranging from only a few thousand to a few million examples. This lack of large-scale, end-to-end multimodal supervision can hinder their ability to tackle complex reasoning tasks. Recent work has shown that MLLMs are vulnerable to distribution shifts, which undermine their reasoning capabilities \citep{wang2024enhancingreasoningabilitymultimodal}, and they can even exhibit catastrophic forgetting and hallucinations \citep{zhai2023investigating}. Furthermore, research indicates that MLLMs struggle with specialized tasks, such as interpreting diagrams, due to limited task-specific training data \citep{joshi2025mmgenenhancingtaskperformance}. Although fine-tuning is an effective method for addressing these challenges, the scarcity or difficulty of obtaining labeled data often makes supervised adaptation impractical in many real-world scenarios, such as medical applications.

To avoid relying on extensive labeled datasets, we introduce a novel Test-Time Warmup (TTW) method which adapts the MLLM on the fly with minimal supervision. Our method, per test instance, leverages data from weakly supervised auxiliary tasks to guide adaptation, allowing the model to refine its representations for complex reasoning without requiring any ground truth annotations. More specifically, we adapt the MLLM individually to each test instance by performing gradient updates on a set of auxiliary tasks derived from that instance. After adaptation, we perform inference on the relevant task instance and discard the updated model weights. 

\begin{figure}[t!]
    \centering    \includegraphics[width=\linewidth]{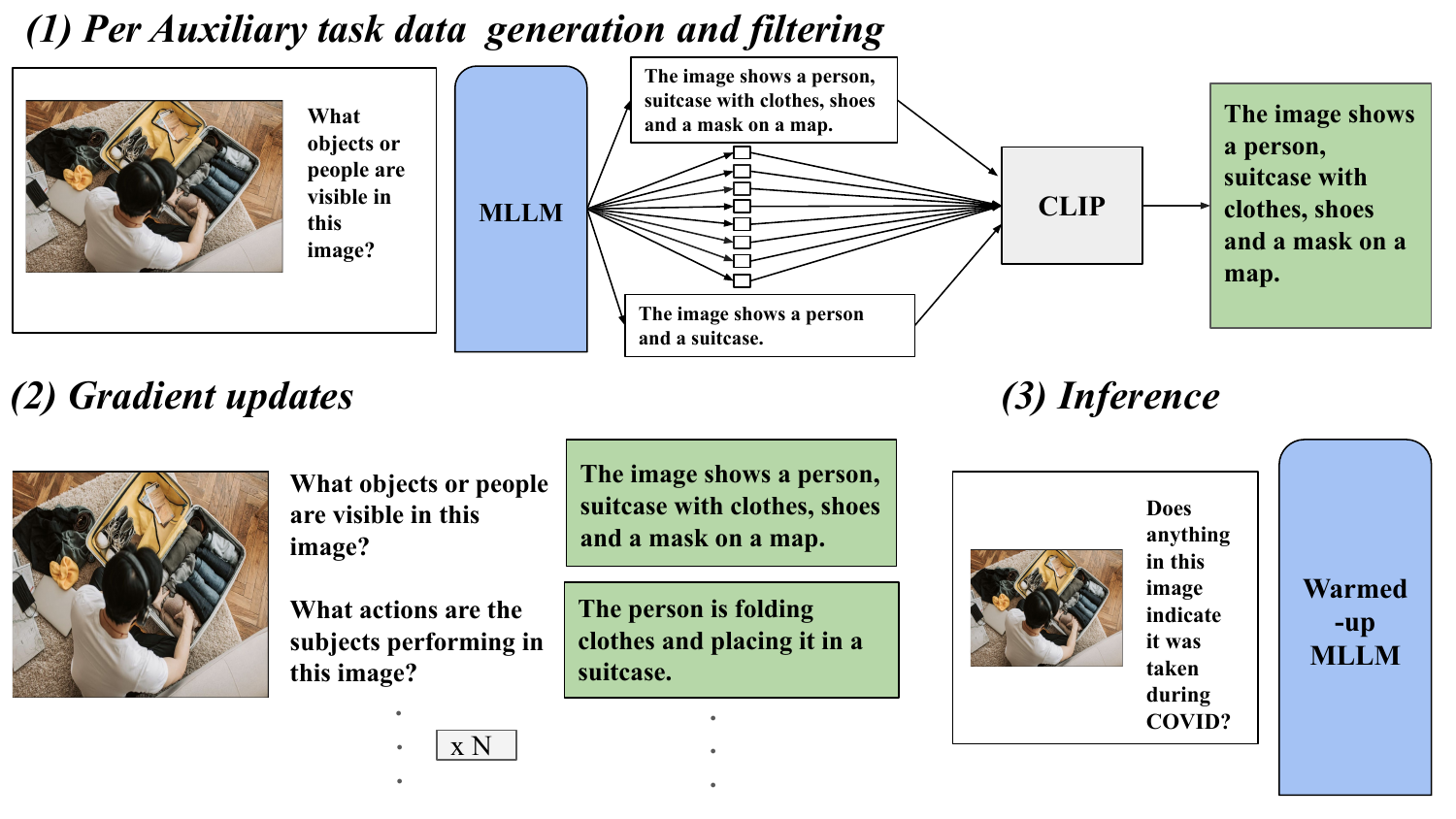}
    \caption{
      An example demonstrating \textbf{Test-Time Warmup} improving a model's attention to detail (a crucial aspect of perceptual reasoning). In step (1), per auxiliary task prompt listed in Figure 2, we generate 10 caption-like outputs. In this example, we show the outputs generated for the object detection auxiliary task. Then, we use the maximum CLIP score to choose one caption. Here, the caption with more details is chosen because it's more aligned with the image. Then in step (2), we perform gradient steps on the chosen caption-like outputs for $N$ auxiliary task prompts ($N=10$ as shown in Figure 2). In doing so, we enforce the model to pay attention to all the objects, including the mask, in the image. For inference, in step (3), the ``warmed-up'' MLLM is better informed to answer the question because it has paid attention to the mask.}
    \label{fig:figure_1_multimodal_work}
\end{figure}

In this work, we focus on evaluating Test-Time Warmup and multimodal large language models (MLLMs) on visual question answering tasks using datasets  such as MMMU \citep{yue2024mmmumassivemultidisciplinemultimodal}, GQA \citep{gqa} and VQA-Rad \citep{narayanan2024freeformmedicalvisual}. Through our evaluation of a popular open-source MLLM, we identify a set of failure modes indicating that these models often struggle when provided with tasks that (1) deviate substantially from the bulk of their training data, or (2) require a highly detailed understanding of the visual content. 

To address these weaknesses, we propose a set of auxiliary tasks to refine image representations and ``warmup'' the model for inference. The auxiliary tasks are designed as input prompts to the target model and are independent of the downstream target tasks. An illustrative analogy is a ``nurse practitioner and doctor'' scenario: the nurse practitioner begins by asking the patient a standardized set of questions, unrelated to any specific ailment, and records the responses for the doctor. With a richer baseline of information about the patient, the doctor can then make a more accurate diagnosis. 

Figure \ref{fig:figure_1_multimodal_work} provides an example of how Test-Time Warmup can prevent an MLLM from making errors. In our experiments, we find that our method outperforms baselines on datasets that focus on perceptual reasoning rather than domain knowledge.  On MMMU, our approach shows gains for plot, chart or tables images: a  form of data that constitutes only a small fraction of the typical training corpus and demands a deeper level of visual comprehension. In the case of GQA, which involves image-question pairs necessitating compositional reasoning, our method boosts performance for scenes that contain many subtle details. Similarly, on VQA-Rad, we see performance improvement on questions that require advanced image understanding. We attribute these gains to our method's ability to foster more nuanced image representations and bring forth existing knowledge in the model rather than provide additional information to the model. 
Our main contributions include the following: 
\begin{itemize}
\item A Test-Time Warmup method for multimodal large language models that uses data from a set of auxiliary tasks to perform gradient steps on the MLLM per test input. After adaptation, we perform inference on the test instance and discard the updated model weights. 
\item Using our method,  we show a relative improvement of 4.03\% on MMMU, 5.28\% on VQA-Rad and 1.63\% on GQA on Llama-Vision-Instruct model \citep{grattafiori2024llama3herdmodels}. 
\end{itemize}

Although we only show results for visual question answering tasks, through this work we motivate the use of our method to perform a diverse set of reasoning tasks. In the case of web agents, for example, performing gradient steps on a fixed set of questions and answers based on a snapshot(s) of a website (e.g., its layout, key features, or common tasks) can help orient the model to the content of that site before it attempts more specialized tasks.
Overall, we offer a lightweight method, without the need for expensive labels, that has the potential to improve MLLM performance on various reasoning tasks.

\begin{figure}[t!]
    \centering
    \includegraphics[width=\linewidth]{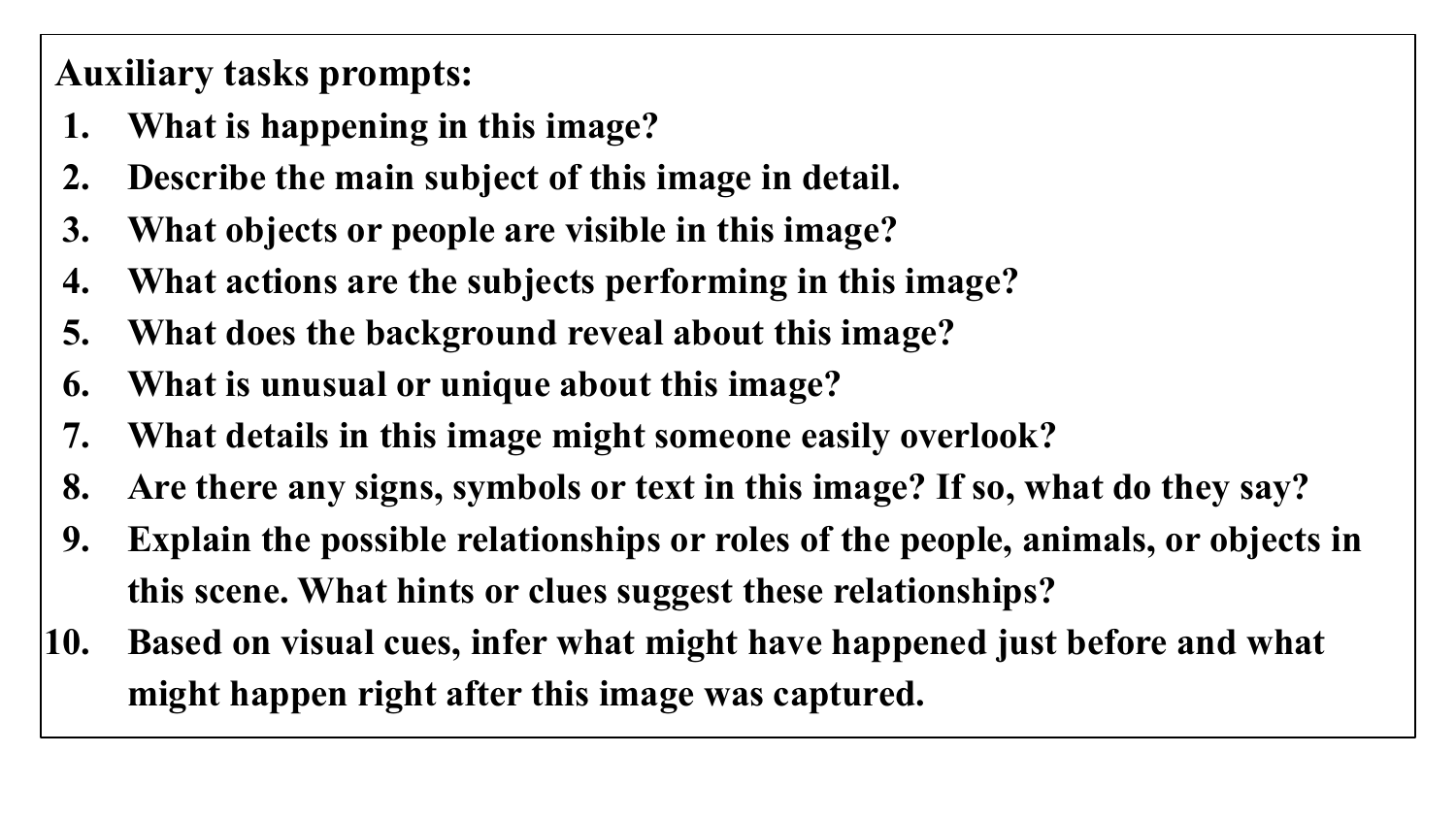}
    \caption{
    Each prompt represents a unique auxiliary task because the prompts are designed to elicit different kinds of information in the image from the MLLM. These auxiliary tasks are not specific to any downstream target task and they aim to refine the target image's representations in the MLLM.
    }
    \label{fig:auxiliary_tasks}
\end{figure}

\section{Related work}

\paragraph{Multimodal large language models}
Recently there has been a lot of interest in developing MLLMs. Well-known open-source models include Llava \citep{liu2023visualinstructiontuning}, Llama-Vision \citep{grattafiori2024llama3herdmodels}, Qwen-VL \citep{bai2023qwenvlversatilevisionlanguagemodel}, and Gemma 3 \citep{gemmateam2025gemma3technicalreport}, and similar model capabilities can also be found in closed-source models such as GPT-4. These models have reasonable performance on tasks similar to their training data but struggle with unseen images or tasks requiring advanced visual reasoning. For example, \citep{wang2024enhancingreasoningabilitymultimodal} highlight that MLLMs are susceptible to distribution shifts, which limit their multimodal reasoning. To address this, they introduce a large-scale multimodal reasoning preference dataset and propose a mixed preference optimization method to improve MLLM reasoning capabilities. This method uses a more implicit approach (preference alignment) while we focus on using auxiliary data that can directly provide context for the test example. Similarly, to improve the performance of MLLMs on specialized tasks with limited prior training, \citep{joshi2025mmgenenhancingtaskperformance} propose a method that utilizes stronger models to generate task-specific synthetic data for fine-tuning MLLMs. Instead of using stronger models, we collect auxiliary task-specific data via the same MLLM, guided by weak supervision from CLIP. This design avoids the heavy reliance on an external, more expensive labeler (human annotators or larger models) and emphasizes the use of internal resources to improve MLLM performance on specialized tasks.

\paragraph{Learning from Auxiliary Tasks}
Researchers have leveraged auxiliary tasks to boost model performance across diverse settings. Chosen as surrogates for the primary objective, these tasks steer learning toward more useful representations. In computer vision, for example, to improve object detection, \citet{Levinshtein2020DATNetDA} train models to learn representations for semantic segmentation and distance transform. One prominent reinforcement-learning example is the UNREAL agent, which augments Atari training with unsupervised pixel-control and reward-prediction tasks that require no extra labels \citep{jaderberg2017reinforcement}. In this work, we weight all auxiliary tasks equally, whereas others, such as \citep{shi_auxiliary_task_reweighting} learn task-specific importance weights. Although our auxiliaries, like most in the literature, are manually designed, automated discovery methods are gaining traction. For instance, \citet{auxiliary_task_discovery_banafsheh} introduce an algorithm that continually generates candidate tasks and keeps those with the highest utility.

\paragraph{Unsupervised Domain Adaptation}
To avoid dependence on large labeled datasets, unsupervised domain adaptation has widely explored using auxiliary tasks across a broad range of domains and applications. For example, \citet{sun2019unsuperviseddomainadaptationselfsupervision} leverage self-supervised auxiliary tasks to align source and target domains, improving generalization in image classification.  \citet{zou2018domainadaptationsemanticsegmentation} generate pseudo-labels from target data and retrain the model to enhance performance in semantic segmentation. \citet{enhancing_unsupervised_vlms} extend this idea to vision-language models (VLMs) by generating domain-aware pseudo-labels for visual feature refinement, enabling better adaptation of models like CLIP for unsupervised domain adaptation (UDA) in classification and detection tasks. Beyond auxiliary-task-based approaches, researchers have also explored clustering and graph-based methods for UDA \citep{learning_sem_reprs_uda, xiao2023spa}. Another prominent line of work involves the use of domain adversarial loss to encourage domain-invariant feature learning and improve model generalization \citep{ganin2016domainadversarialtrainingneuralnetworks, Cai_2020}.

\paragraph{Test-Time Adaptation}
An increasingly popular subfield within domain adaptation is test-time adaptation (TTA), which focuses on adapting models during inference time. test-time adaptation is particularly powerful when each test sample might come from a very different distribution, one that is difficult to handle via traditional fine-tuning. Tasks such as Abstract Reasoning Challenge (\citep{akyürek2024surprisingeffectivenesstesttimetraining}) and Massive Multi-discipline Multimodal Understanding and Reasoning \citep{yue2024mmmumassivemultidisciplinemultimodal} are particularly well-suited for test-time adaptation because each test sample or a subset of samples often come from disparate distributions. Researchers first began to look at test-time training for image classification. \citet{test_time_training_for_ood} introduced a test-time training method that uses image rotation prediction as the auxiliary task. \citet{wang2021tent} proposed test-time entropy minimization to reduce generalization error for image classification. \citet{khurana2022sitasingleimagetesttime} computed normalization statistics of a single unseen test-time image to perform image classification and segmentation.

Our work is also inspired by recent advances in using test-time adaptation for the Abstraction and Reasoning Challenge (ARC\citep{clark2018thinksolvedquestionanswering}). In this perceptual challenge, each example differs significantly from every other, making it difficult for traditional approaches to learn a unified representation across all possible inputs. Similarly, images (especially from different domains) can look different from each other which motivates the use of test-time adaptation to improve perceptual reasoning. \citet{akyürek2024surprisingeffectivenesstesttimetraining} achieve a six-fold performance improvement on ARC by applying test-time training. They enhance performance by generating auxiliary task labels and training task-specific LoRA adapters, which are discarded after inference on the target task. \citet{cole2025dontthrowbabybathwater} 
similarly used test-time updates to enforce the model to perceive the ARC examples in the correct way among the the near-infinite possible image abstractions for an ARC task. They encode each ARC task (and examples) as text-sequence to train a LongT5 encoder-decoder model at test-time.

\paragraph{In-context Learning}
While test-time Training explicitly updates model parameters, In-Context Learning (ICL) instead conditions them on input at inference time without any gradient-based updates. This approach has demonstrated strong performance on a range of new tasks. For example, \citet{icl_auxtask} leverage auxiliary information in-context to improve short-text representations. Although prior work \citep{akyrek2023what} suggests that ICL might function as an implicit form of machine learning simulation, empirical results show that ICL with language models often does not align with standard machine learning algorithms \citep{zhao2024probing,min-etal-2022-rethinking}. In addition, ICL can struggle with unfamiliar tasks “out-of-the-box,” as evidenced by the poor performance of large language models on datasets such as ARC \citep{opiełka2025analogicalreasoninginsidelarge,boberirizar2024neuralnetworksabstractionreasoning}.
\section{Test-Time Warmup}

In this section, we describe our algorithm for Test-Time Warmup for MLLMs.  See Figure~\ref{fig:figure_1_multimodal_work} for an example application of our method.

\subsection{Auxiliary tasks}
Ideally, generating auxiliary task information should be lightweight, requiring minimal additional computation. Beyond being cost-effective, our goal is to produce informative auxiliary data that effectively enhances multimodal reasoning capabilities.

To achieve this, we construct prompts (see Figure~\ref{fig:auxiliary_tasks}) that elicit specific types of visual information from the MLLM that might be useful context for performing more difficult tasks later. For instance, the prompt \emph{"Are there any signs, symbols, or text in this image? If so, what do they say?"} aims to improve text recognition within images. Meanwhile, \emph{"What objects or people are visible in this image?"} functions like an object detection task. Another example, \emph{"Based on visual cues, infer what might have happened just before and what might happen right after this image was captured,"} encourages temporal and contextual reasoning, a valuable skill for certain types of reasoning tasks. 

Each auxiliary task yields caption-like outputs. We expect these caption-like outputs to be reasonable because captioning tasks comprise a significant portion of the MLLM training mix, and thus the model may have more general abilities in captioning than other, more specialized tasks. Furthermore, CLIP models can be used to weakly supervise caption quality, as this directly aligns with their large-scale training objective.

\paragraph{Data generation} To produce the auxiliary data for each test image, we prompt the target MLLM for 10 distinct “caption-like” responses per auxiliary task. We use vLLM for faster generation and set the temperature to $0.75$ to introduce variation in the responses. This process yields a set of diverse candidate outputs for each prompt and image.

\paragraph{Data filtering} Next, we apply a filtering step using CLIP to ensure only the most relevant and visually faithful caption is retained. Specifically, for each auxiliary task, after we generate 10 captions, we select the one with the highest CLIP score. This yields a set $N$ captions, where $N$ is the number of auxiliary tasks (we choose $N = 10$).  This weakly supervised filtering ensures that the retained set of captions best represent the image content.

\subsection{Gradient updates} Having generated and filtered the auxiliary data, we adapt the MLLM individually to each test image by performing gradient-based updates on its caption-like outputs. We freeze the vision encoder and train only the LLM and connector components using a cross-entropy loss. (We also considered other objectives, including GRPO; see \emph{Future Work} for further discussion.) These gradient steps effectively ``warmup'' the model to the specific visual details of the test instance before tackling the downstream task.

\subsection{Inference at test time} Once the model is adapted for a given image, we perform inference on the downstream task and then discard the updated weights. We repeat the same data generation, filtering, gradient adaptation, and inference procedure for the next test image. This per-instance adaptation strategy equips the model with a detailed understanding of each image at inference time without permanently altering the global model parameters.
\section{Experiments}

We conduct our experiments using the Llama-3.2-11B-Vision-Instruct model and 4 different datasets (see below for details of each). Throughout all experiments, the MLLM vision encoder remains frozen, and only the connector and LLM components are trained at test time. For each dataset, we randomly sample 500 images for our test set. We set the learning rate to 1e-6, use the AdamW optimizer \cite{loshchilov2019decoupledweightdecayregularization}, a batch size of 5, and train for 2 epochs. To ensure the model does not overfit to a specific sequence of auxiliary data, we shuffle the auxiliary examples for each image at the start of every epoch.
We use CLIP-ViT to weakly supervise auxiliary data for MMMU, GQA and VQA. For VQA-Rad, which is a medical dataset, we use BiomedCLIP \citep{zhang2024biomedclipmultimodalbiomedicalfoundation}. 

\subsection{Datasets}
We evaluate our method on the following datasets:

\noindent\textbf{MMMU} \citep{yue2024mmmumassivemultidisciplinemultimodal} is designed to test multimodal understanding and reasoning capabilities at an expert level. It contains 30 subjects, including arts, science, medicine, and law, and features expert-authored multiple-choice questions that often require detailed analysis of diagrams, plots, charts, and medical images.

\noindent\textbf{GQA} \citep{gqa} is a large-scale VQA benchmark derived from Visual Genome, containing 22M questions spanning 113K images. It emphasizes compositional reasoning and scene understanding, with questions grounded in a scene graph to enable fine-grained evaluation of visual reasoning.

\noindent\textbf{VQA-RAD} \citep{narayanan2024freeformmedicalvisual} 
is a medical VQA dataset with over 3,500 clinician-authored question-answer pairs covering 315 radiology images. It focuses on clinically relevant reasoning (e.g., identifying anatomical structures or abnormalities) and includes both open- and closed-ended questions, making it an essential benchmark for medical AI systems.

\noindent\textbf{VQA} \citep{vqav2}
is a large-scale open-ended VQA dataset consisting of 250,000 images paired with over 760,000 questions. It covers a broad range of reasoning types, such as object recognition, counting, and commonsense inference, challenging models to jointly understand vision and language.

\subsection{Baselines}
\paragraph{Base MLLM} First, we consider a simple baseline. We use the huggingface open-source weights of Llama-3.2-11B-Vision-Instruct to evaluate on the sample set of 500 images for each dataset. 

\paragraph{In-Context Learning + Our auxiliary data} To demonstrate the benefits of performing test-time training on auxiliary task prompts (rather than simply providing them as in-context information for the target task), we consider the baseline where we only supply the auxiliary data (generated using the process shown in Figure \ref{fig:figure_1_multimodal_work})
within the test-time prompt and without any gradient updates. The detailed prompt used for this experiment is provided in the Appendix.

\subsection{Evaluation} 
We evaluate models using response accuracy. For GQA, VQA-Rad, and VQAv2, a response is considered correct if it contains the ground-truth answer. Additionally, for VQAv2, which includes multiple annotations, we apply the standard soft-scoring scheme that assigns higher weight to answers occurring more frequently in the annotation set. For MMMU, we report accuracy using the official evaluation code \citep{yue2024mmmumassivemultidisciplinemultimodal}. \footnote{We make a single change in the official code for reproducibility: for multiple-choice questions where no option can be parsed from the model’s response, we deterministically select the first option rather than a uniform random choice over the listed options.} To ensure consistency, for a given dataset all baselines, methods, and ablations use the same evaluation strategies. The exact prompts and sampling parameters are provided in the appendix.

\section{Results and Discussion}

In this section, we empirically study our method's ability to improve MLLM reasoning capabilities. We also perform ablations on the VQA-Rad dataset to understand the significance of each component in our method.

\begin{table}[h]
\centering
\begin{tabular}{lccc}
\toprule
\textbf{Dataset} & \textbf{Base MLLM} & \textbf{ICL + Our auxiliary data} & \textbf{TTW [Ours]} \\
\midrule
GQA      & 61.2  & \textbf{62.8} ($\Delta$ 2.61\%) & 62.2  ($\Delta$ 1.63\%) \\
MMMU     & 44.6 &  44.6 ($\Delta$ 0\%) & \textbf{46.4}  ($\Delta$ 4.03\%) \\
VQA-Rad  & 49.2  & 48.6  ($\Delta$ -1.22\%) & \textbf{51.8}  ($\Delta$ 5.28\%) \\
VQAv2    & 73.64 & 72.3 ($\Delta$ -1.82\%)  & \textbf{73.84} ($\Delta$ 0.27\%) \\
\bottomrule
\end{tabular}
\caption{Comparison of accuracy on Base MLLM (LLama-3.2-11B-vision-instruct), ICL (which includes auxiliary data supervised by CLIP from our method), our Test-Time Warmup method on datasets GQA, MMMU, VQA-Rad and VQAv2. In parenthesis we show the relative improvement in accuracy of Test Time Warmup compared to the base MLLM. Our model mainly shows improvement on datasets that require more advanced perceptual reasoning.}
\label{tab:main_results}
\end{table}

\subsection{Overall performance gains from Test-Time Warmup}

Our method delivers its strongest gains on benchmarks that emphasize perceptual reasoning. As Table \ref{tab:main_results} shows, Test-Time Warmup increases performance on MMMU, GQA, and VQA-Rad, all tasks that demand detailed visual understanding. In contrast, it provides only a modest improvement on VQAv2, a benchmark that leans more on general world knowledge than on perceptual cues. On GQA, ICL+auxiliary data edges out TTW, likely because the model has already seen many natural images and can exploit in‑context learning without additional gradient updates. In contrast, MMMU and VQA‑Rad benefit far more from TTW, probably because the model’s pre‑training included few plot‑style or radiology images, making in-context learning alone insufficient.

For MMMU, GQA, and VQAv2, the reported accuracies of the base Llama-3.2-11B-Vision-Instruct model (Table \ref{tab:main_results}) are obtained using basic inference techniques such as direct prompting without beam search. On MMMU, subject-specific chain-of-thought prompting for eg., can add a few percentage points. Test-Time Warmup is orthogonal to both prompting and other advanced sampling strategies and combining them can yield even greater gains.
Further, although specialized medical models like LLaVA-Med achieve far higher VQA-Rad accuracies, we demonstrate that a Test-Time Warmup on a general-purpose model can still realize measurable accuracy gains by using an inexpensive medical expert such as BioMedCLIP to supervise the auxiliary data.

A key factor in attaining the results shown in Table \ref{tab:main_results} was the carefully curated set of auxiliary prompts in Figure \ref{fig:auxiliary_tasks}, chosen to balance diversity with utility. Preliminary experiments drew ten prompts at random from a pool of fifty, but the extra variety often produced noisy, less helpful responses. We further observed that open‑ended prompts (in Figure \ref{fig:auxiliary_tasks}) were superior to constrained ones, whether limited to a specific sentence count or generated from object‑detection templates (e.g., “Is object A left of object B?”). These constrained prompts offered little benefit on the higher‑level reasoning tasks we target.

Figures \ref{fig:gqa_examples}, \ref{fig:vqa-rad_examples} and  \ref{fig:mmmu_1}, \ref{fig:mmmu_2} (in appendix) present qualitative examples from the GQA, VQA-Rad, and MMMU datasets. On MMMU, improvements were most apparent on plot- or chart-based images, which are underrepresented in typical MLLM training data and demand deeper visual comprehension. For GQA, gains were observed on examples requiring compositional reasoning and attention to subtle image details. Similarly, on VQA-Rad, the model performed better on questions involving more advanced image understanding. Across these examples, two distinct patterns emerge: (1) auxiliary task responses that explicitly surface information that directly helps answer the test question, and (2) auxiliary task responses that provide additional context or a fresh perspective on the image.

In general, we find that TTW simply “surfaces” knowledge the MLLM already holds. We sample candidate auxiliary responses, from the MLLM,  at temperature 0.75 to encourage diversity, rank them with CLIP, and then take gradient steps that increases the log-likelihood of the top-ranked response. This nudges the model toward information it previously under-utilized. For instance, in the far-right GQA example in Figure \ref{fig:gqa_examples}, TTW enforces the model to note the white bench—information it ignored without warmup and thus answered “headboard” incorrectly. Because CLIP can only elevate a response the model has already produced, every improvement reflects latent knowledge that TTW helps deploy.

Feeding the model a diverse set of auxiliary prompt responses improves its image-text alignment and equips it for more challenging reasoning tasks. Each auxiliary prompt encourages the model to view an image from a new angle. For instance, in the far-right VQA-Rad example in Figure \ref{fig:vqa-rad_examples}, the auxiliary task prompt to identify anything unusual steers the model toward relevant anomalies, enabling the correct answer. By mitigating gaps in under-represented domains in  MLLM training data, evidenced by our gains on MMMU and VQA-Rad, a diverse set of auxiliary data cultivates more robust image representations.

In the appendix, we also show examples of failure modes of our method. 

\begin{figure}[t!]
    \centering    \includegraphics[width=\linewidth]{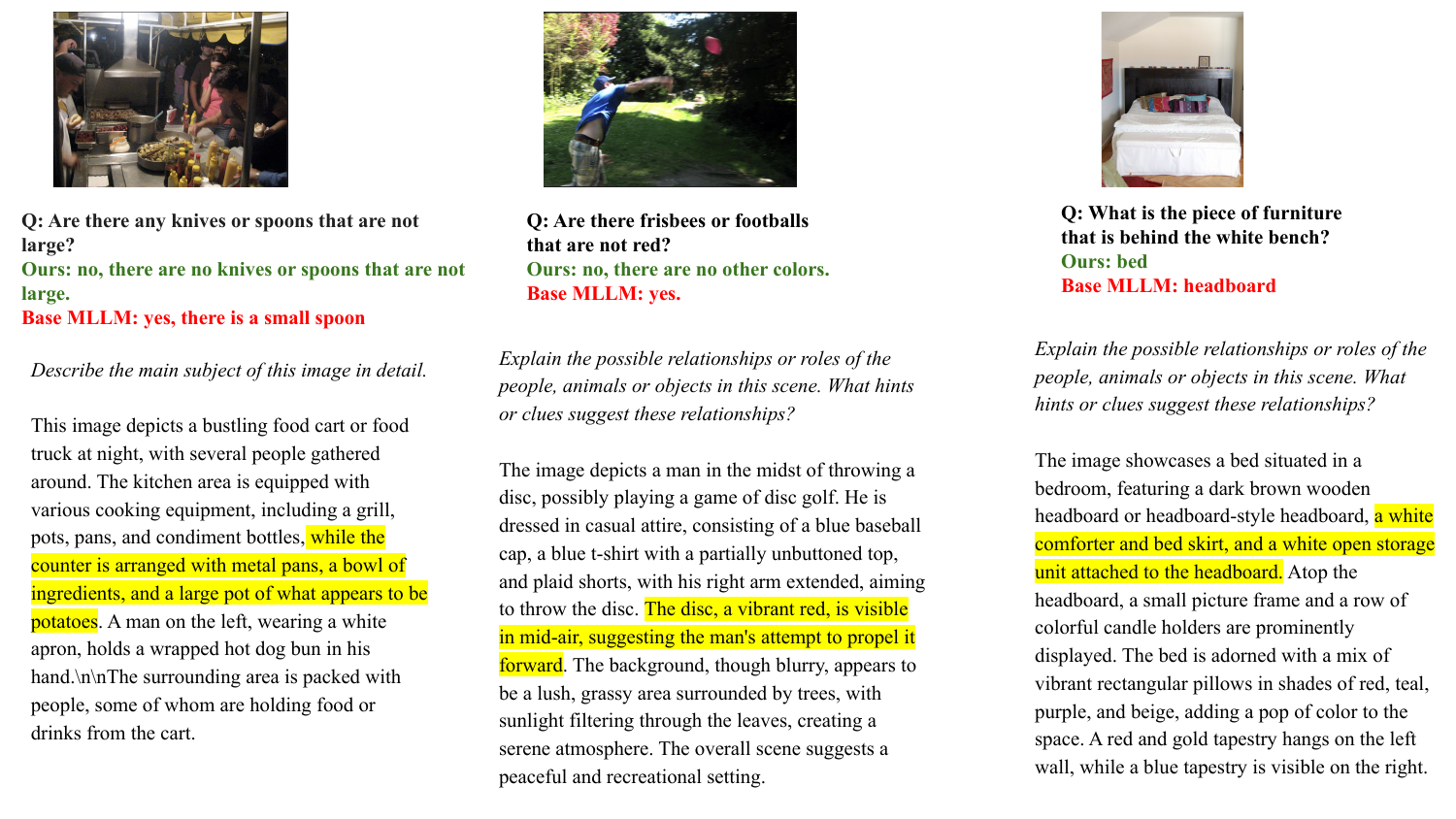}
    \caption{
    Examples in GQA of where our "warmed up" MLLM answers correctly while the base MLLM fails. Below each example we show the auxiliary prompt response that was the most relevant to the question and potentially aided the model for better reasoning. For the left‑most image, crowded with objects and requiring fine‑grained attention, the auxiliary response adds scene context that helps the model infer there are no small spoons. In the centre and right‑most images, the auxiliary response explicitly provides information needed to answer the question, directly steering the model toward the correct answer.
    }
    \label{fig:gqa_examples}
\end{figure}

\begin{figure}[t!]
    \centering    \includegraphics[width=\linewidth]{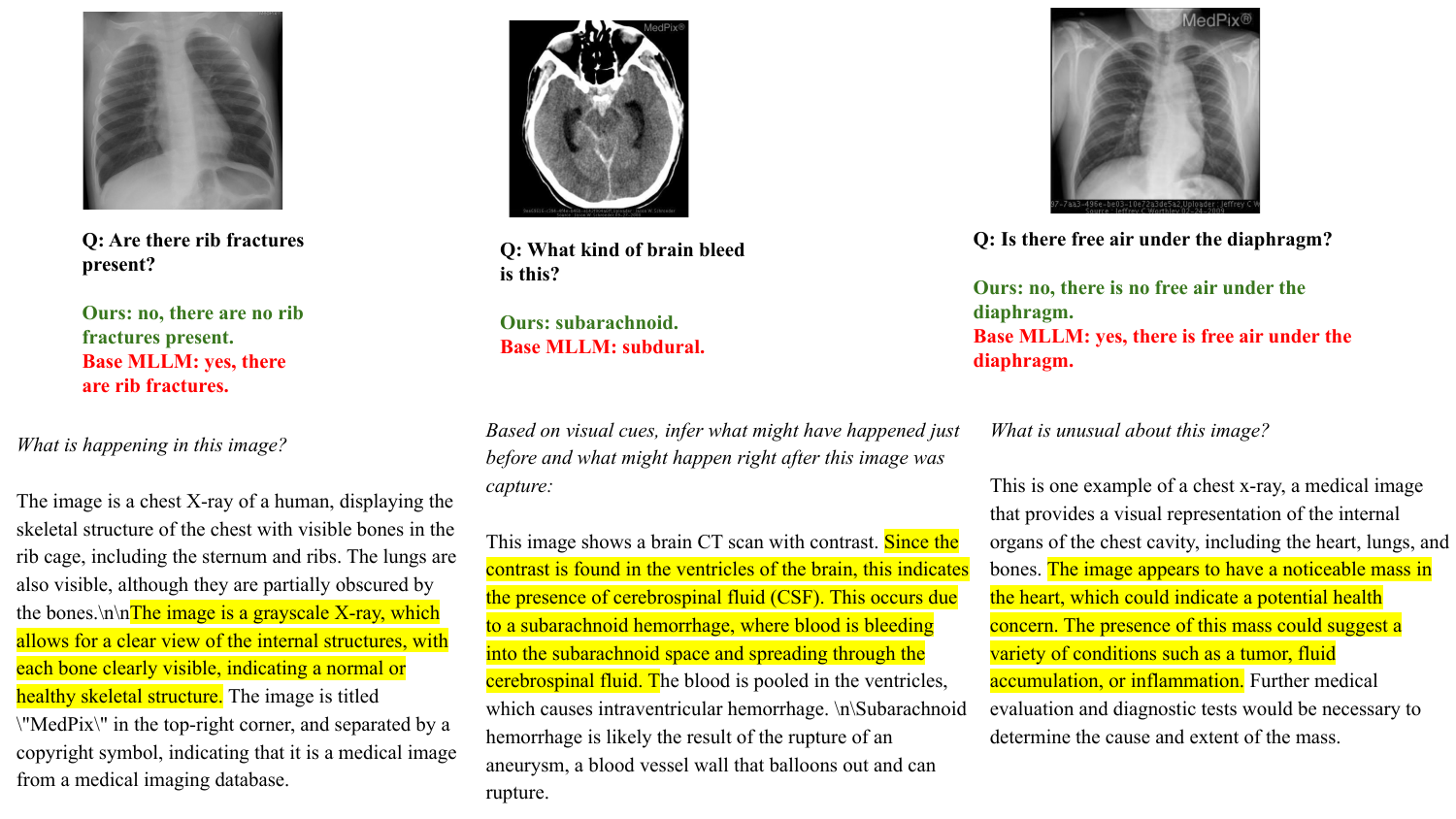}
    \caption{
    Examples in VQA-Rad of where our "warmed up" MLLM answers correctly while the base MLLM fails. Below each example we show the auxiliary prompt response that was the most relevant to the question and potentially aided the model for better reasoning. For the left‑most and centre images, the auxiliary response delivers information directly needed to answer the test question. Although the auxiliary response for the far‑right image is not directly tied to the question, including it in Test-Time Warmup compels the model to examine the image carefully beforehand, helping it avoid hallucinating findings that are not present.
    }
    \label{fig:vqa-rad_examples}
\end{figure}

\subsection{Ablations}
We perform CLIP-related ablations on the VQA-Rad dataset. We chose this dataset because we hypothesized that a medical expert, BiomedCLIP, supervising a non-medical model, Llama-3.2-11B-Vision-Instruct, would provide the most significant boost in performance. 

\subsubsection{Auxiliary Task Data Generation without supervision} To isolate the contribution of weak supervision from BiomedCLIP for VQA-Rad, we run our method without it. Specifically, we generate a single caption per prompt (instead of 10) during auxiliary data generation and perform no filtering. This ablation clarifies the performance gains attributable to BiomedCLIP-based selection of the most relevant caption-like output.  As shown in Table \ref{tab:ablations_biomed_clip}, we see that BiomedCLIP supervision does improve model performance.

\subsubsection{Auxiliary task data generation with a poor reward model} In the auxiliary data filtering stage, after generating 10 prompts per auxiliary task, we choose the worst generation (the lowest CLIP score we get from BiomedCLIP) to train our model on. This helps understand how a poor reward model can affect test-Time Warmup. In Table \ref{tab:ablations_biomed_clip}, we see that there's a small difference in doing auxiliary data generation without BiomedCLIP or using the inverse of BiomedCLIP. This could possibly be due to the fact that the worst generation is bounded by the target MLLM. Additionally, the inverse of BiomedCLIP performing better than without BiomedCLIP can be due to a small amount of noise in BiomedCLIP scores.

\begin{table}[h]
\centering
\resizebox{\textwidth}{!}{%
\begin{tabular}{lccccc}
\toprule
\textbf{Dataset} & \textbf{Base MLLM} & \textbf{Ours [TTW]} & \textbf{Ours w/o BiomedCLIP} & \textbf{Ours w/ Inverse of BiomedCLIP} \\
\midrule
VQA-Rad  & 49.2  & \textbf{51.8} & 48.8 & 49.6 \\
\bottomrule
\end{tabular}
}
\caption{Comparison of accuracy of the Base MLLM (Llama-3.2-11B-Vision-Instruct) and ablations of BiomedCLIP on the VQA-Rad dataset. BiomedCLIP significantly improves  accuracy on VQA-Rad.}
\label{tab:ablations_biomed_clip}
\end{table}

\section{Limitations and Future work}

To improve the performance and  efficiency of our method and to strengthen the claims we make in this work, we consider the following directions. 

\paragraph{Models} 
We tested our method on Qwen-VL-Chat \citep{bai2023qwenvl} and Gemma-12B-it \citep{gemmateam2025gemma3technicalreport}, but observed negligible improvements (less than $\Delta0.2$ in accuracy). This highlights an important limitation, while also underscoring the need to further establish the method’s model-agnostic nature—so that, like chain-of-thought prompting, it can serve as a general framework for enhancing reasoning across diverse models.

\paragraph{LoRA}
One of the main limitations of our work is that it is computationally expensive. Instead of training the entire connector and LLM, it would be helpful to use LoRA \citep{hu2021loralowrankadaptationlarge} adapters. LoRA is especially helpful in settings where there is limited data. It allows efficient adaptation while maintaining the model's general capabilities.

\paragraph{GRPO + online auxiliary task generation}
In our current approach, we use CLIP to filter auxiliary task data. Instead of filtering and discarding potentially useful information, a more effective method could be to leverage GRPO. \citet{deepseekai2025deepseekr1incentivizingreasoningcapability} optimize a preference using group scores of outputs generated from a single prompt. By using GRPO instead of filtering, we would avoid discarding potentially useful information. CLIP can act as the reward model. For each auxiliary task prompt, we would obtain CLIP scores for n captions and compute the GRPO loss to update the model accordingly. To obtain the most benefits from using GRPO, it could be useful to generate new responses for the auxiliary prompts after each epoch and create a self-improvement loop. 

\paragraph{Process supervision metric} We show that our method can improve the accuracy of an MLLM on benchmarks that test perceptual reasoning capabilities. To confirm that our method directly improves reasoning, it would be helpful to develop a process supervision metric that measures a model’s reasoning steps before it arrives at a solution. A possible approach to do this is to use a stronger MLLM to judge the reasoning steps of the target MLLM. 

\begin{figure}[t!]
    \centering
    \includegraphics[width=\linewidth]{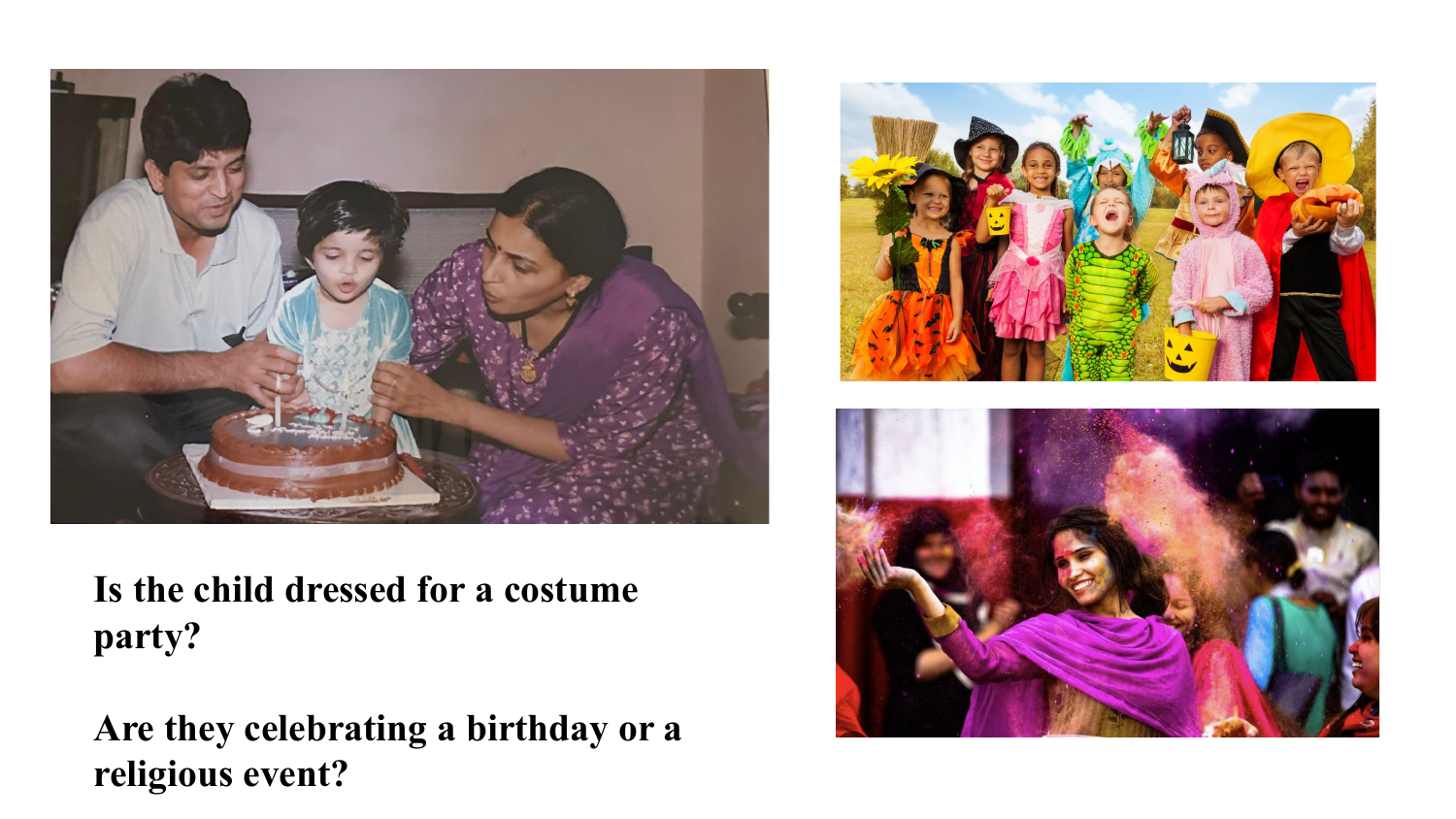}
    \caption{
     This figure motivates using Test-Time Warmup for AI safety in MLLMs. Both the prompts shown in this figure are only offensive conditional on the image on the left because they prompt the model to make assumptions about traditional attire. For the image with children in Halloween costumes, the first question would be appropriate and for the image in the bottom right the second question would be appropriate because they are celebrating holi, a religious event.  }
    \label{fig:safety_alignment_test_time}
\end{figure}

\paragraph{Diverse set of reasoning tasks} 
As mentioned before, we built Test-Time Warmup to adapt MLLMs for a diverse set of reasoning tasks. In this paper, we only show results for visual question answering tasks but in the future we believe an exciting direction would be to show results for web agents and other tasks that can benefit from our method. 

\paragraph{Optimal set of auxiliary tasks}
We selected a small set of auxiliary task prompts that best mitigated the failure modes we observed. Future work could replace this manual selection with data-driven strategies for discovering the most effective auxiliary tasks. For example, \citep{ruis2025proceduralknowledgepretrainingdrives} show that, for reasoning questions, training examples rich in procedural knowledge (e.g., step‑by‑step formulas or code) are highly influential. A similar empirical study focused on perceptual reasoning could reveal insights to steer auxiliary‑task design.

\paragraph{Test-Time Warmup towards AI Safety}
Most LLM red‑teaming efforts build large, diverse prompt sets to test whether a model can refuse unsafe requests and to fine‑tune its safety alignment \citep{perez2022redteaminglanguagemodels, samvelyan2024rainbowteamingopenendedgeneration}. \citep{zollo2024effectivediscriminationtestinggenerative} show that such static red team datasets cannot anticipate threats that arise in complex, real-world interactions. The gap widens in multimodal settings, where a prompt may be harmful only when paired with a specific image. Our approach provides a foundation towards safety at test-time. For each incoming image, we can first use a strong MLLM (similar to what \citep{perez2022redteaminglanguagemodels} do with LLMs) to generate potential “attack prompts” (instead of using our current static set of auxiliary task prompts), then immediately adapt the target MLLM so it refuses or safely answers those prompts. Our Test-Time warmup approach can greatly expand coverage of image‑conditioned threats. Figure \ref{fig:safety_alignment_test_time} illustrates a case where a prompt is unsafe only in the presence of the accompanying image.

\section*{Acknowledgments}
We thank Sruthi Sudhakar, Todd Morrill, Prof. Kaveri Thakoor, Prof. Kent Quanrud and Prof. Carl Vondrick for very helpful feedback on this work. We also thank the Google Cyber Research Program and ONR (Award N00014-23-1-2436) for their generous support. This work is supported by the funds provided by the National Science Foundation and by DoD OUSD (R\&E) under Cooperative Agreement PHY-2229929 (The NSF AI Institute for Artificial and Natural Intelligence).


\bibliography{references}

\clearpage
\appendix
\section{Additional Experiment Details and Results}
Here, we discuss failure modes of our method and include more experiment details . Our code is available at \url{https://github.com/nrajanee/test-time-warmup-mllms}.

\subsection{Failure modes of our method} 

By generating the auxiliary warm-up data with the target MLLM, we remove the need for labeled datasets, though this introduces a key failure mode. The MLLM may hallucinate, and although CLIP supervises these generations to curb hallucinations, subtle errors can evade detection. If CLIP selects such a hallucinated output, the ensuing answer may also be incorrect.

\begin{figure}[t!]
    \centering    \includegraphics[width=\linewidth]{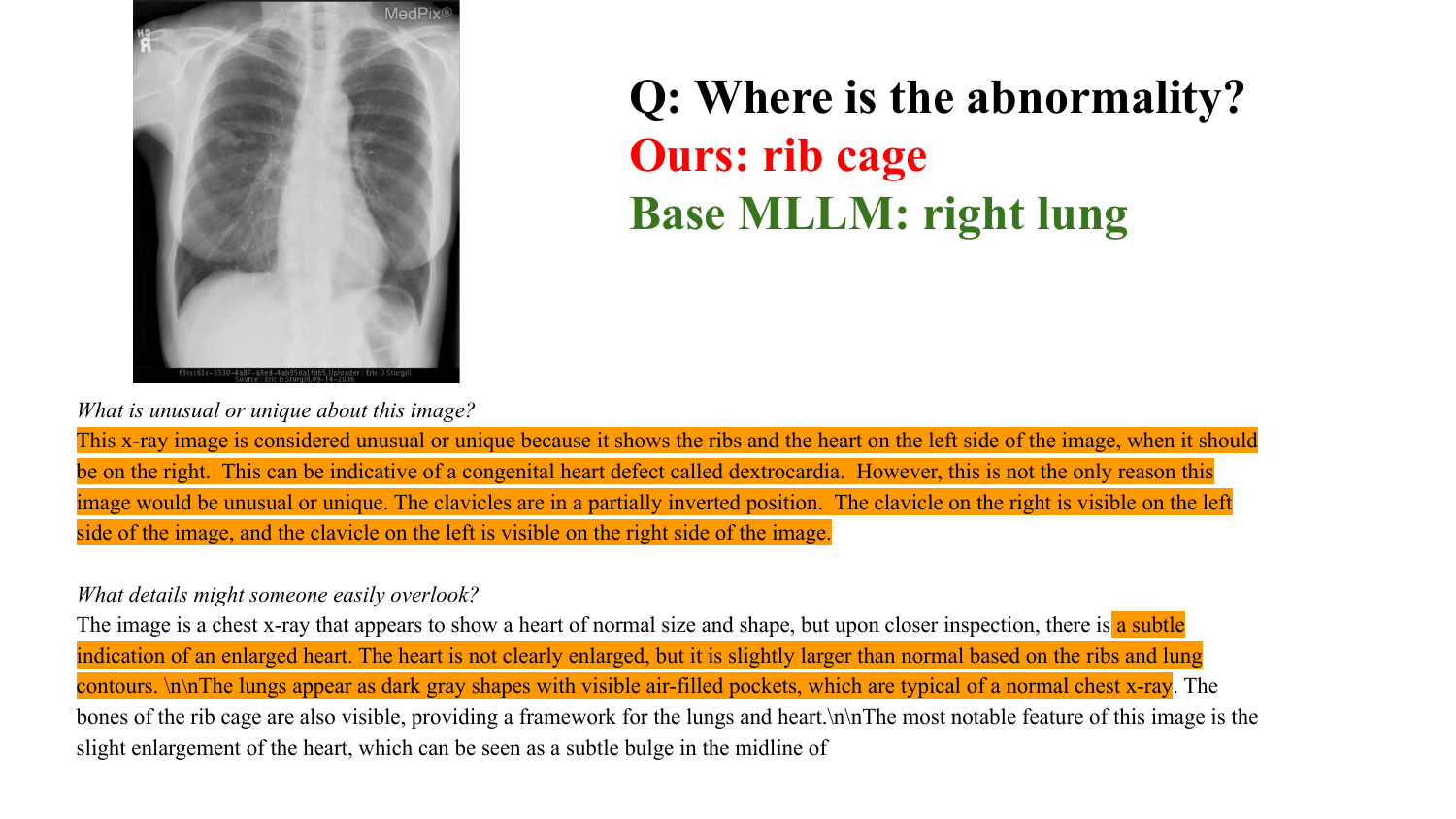}
    \caption{
     In this example from VQA-Rad, where the warmed-up MLLM responds incorrectly while the base model answers correctly. Both of these auxiliary task outputs contain hallucinations about the ribs, heart and clavicle which likely led to the model answering incorrectly.  
     }
    \label{fig:vqa-rad_failed}
\end{figure}

\begin{figure}[t!]
    \centering    \includegraphics[width=\linewidth]{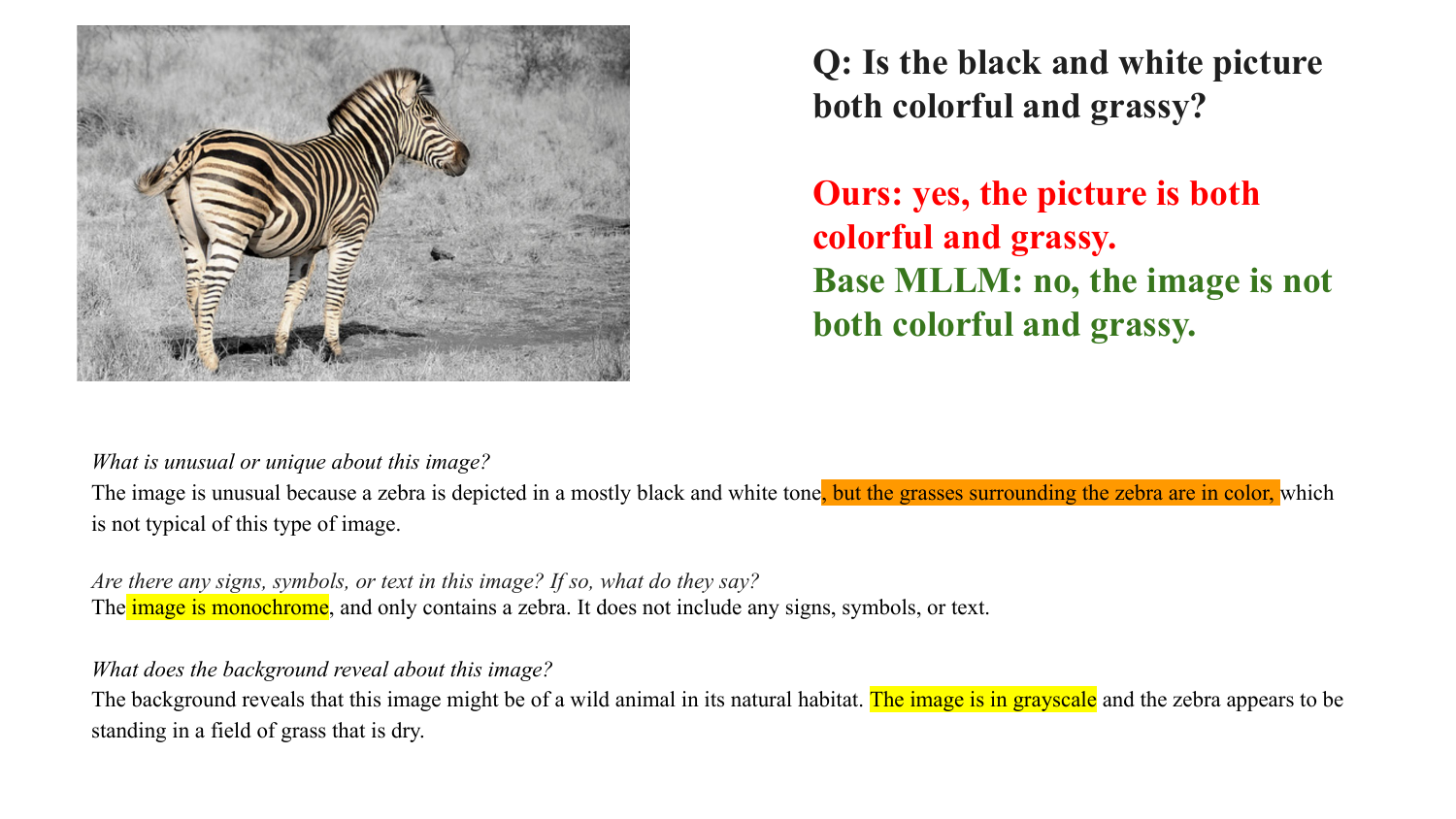}
    \caption{
    In this GQA example, the warmed‑up MLLM responds incorrectly while the base model answers correctly. Despite weighting all auxiliary task outputs equally and reshuffling them before each epoch, a single hallucinated output describing colorful grass, in contrast to the others’ grayscale description, still steers the model to the wrong answer.
     }
    \label{fig:gqa_failed}
\end{figure}

\begin{figure}[t!]
    \centering    \includegraphics[width=\linewidth]{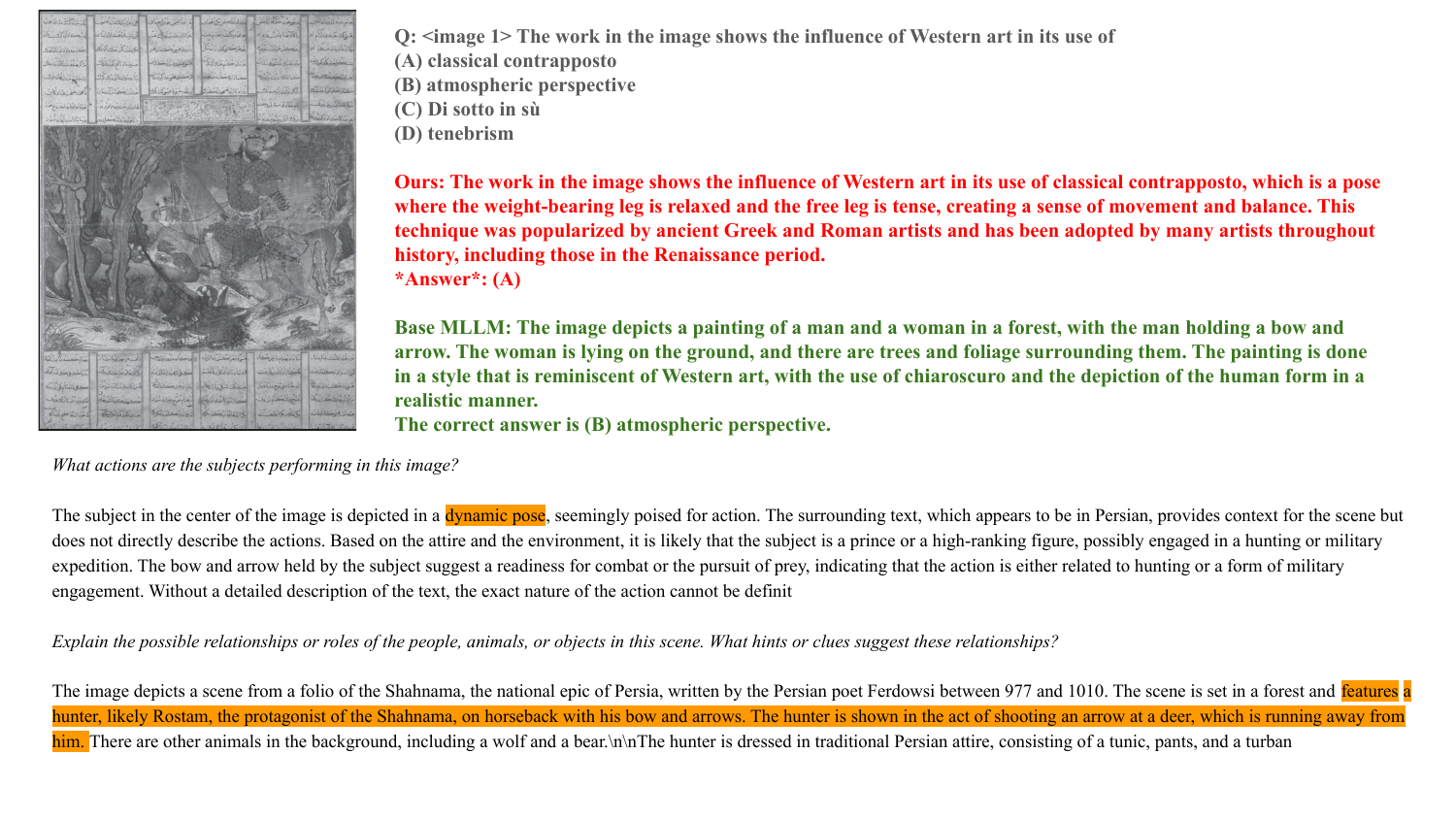}
    \caption{
        In this MMMU example, the warmed-up MLLM responds incorrectly while the base model answers correctly. Both of these auxiliary task outputs emphasized the pose and action of the figure rather than the scenery, which likely led the model to answer incorrectly.
     }
    \label{fig:mmmu_failed_1}
\end{figure}

In Figures \ref{fig:vqa-rad_failed}, \ref{fig:gqa_failed} and \ref{fig:mmmu_failed_1} we show examples of this failure mode in MMMU, GQA and VQA-Rad. 

\begin{figure}[t!]
    \centering    \includegraphics[width=\linewidth]{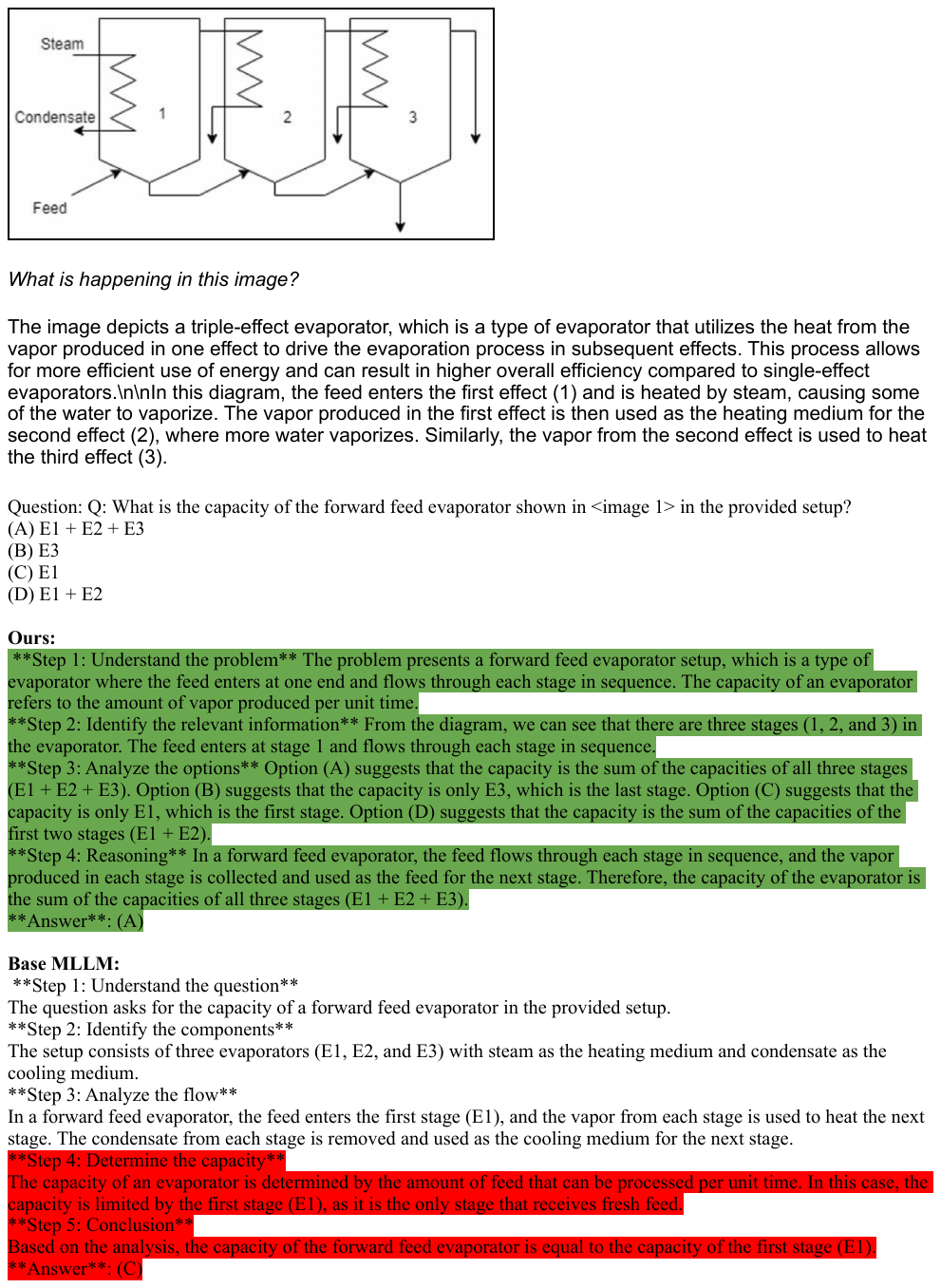}
    \caption{
     Example in MMMU of where our "warmed up" MLLM answers correctly while the base MLLM fails. On the left we show the auxiliary prompt response that was the most relevant to the question and potentially aided the model for better reasoning. On the right, we show the answer response with correct reasoning highlighted in red for incorrect reasoning. Here the auxiliary response explicitly provides information relevant to the question, which demonstrates how these auxiliary task prompts can help the model focus on the critical details needed to reach the right conclusion. 
    }
    \label{fig:mmmu_1} 
\end{figure}

\begin{figure}[t!]
    \centering    \includegraphics[width=\linewidth]{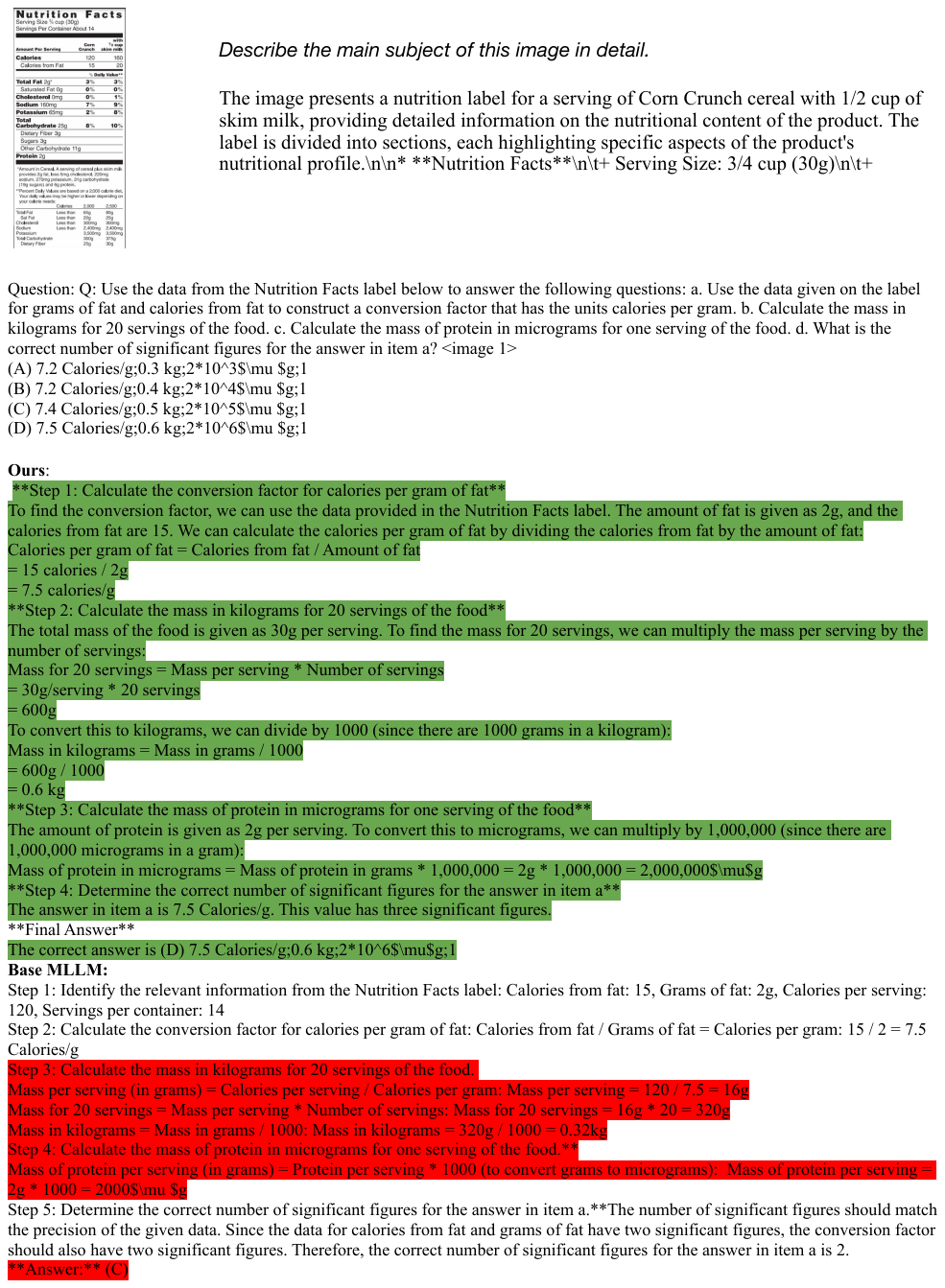}
    \caption{
     Example in MMMU of where our "warmed up" MLLM answers correctly while the base MLLM fails. On the left we show the auxiliary prompt response that was the most relevant to the question and potentially aided the model for better reasoning. On the right, we show the answer response with correct reasoning highlighted in red for incorrect reasoning. Although the auxiliary response does not directly answer the test question, warming the model up with contextual information about the nutritional information guides its attention to the key details, enabling it to arrive at the correct answer.
     }
    \label{fig:mmmu_2}
\end{figure}

\subsection{Evaluation}

\paragraph{Metric}
We evaluate models using zero-shot accuracy. For GQA, VQA-Rad, and VQAv2, a response is
considered correct if it contains the ground-truth answer. Additionally, for VQAv2, which includes
multiple annotations, we apply the standard soft-scoring scheme that assigns higher weight to answers
occurring more frequently in the annotation set. For MMMU, we report accuracy using the official
evaluation code Yue et al. (2024) 1. To ensure consistency, all baselines, methods and ablations use
the same evaluation strategies for each dataset. The exact prompts and sampling parameters are
provided in the appendix. 

\paragraph{GQA, VQA-Rad, VQAV2} 
For evaluating Llama-3.2-11B-Vision-Instruct on these datasets we use the prompts shown in Table \ref{tab:prompts}. To generate the responses from the model, we use greedy sampling and we set the sampling parameter max\_new\_tokens = 128. 

\paragraph{MMMU} 
For evaluating LLama-vision-instruct on MMMU we use the prompts shown in Table \ref{tab:prompts}. To generate the responses from the model, we set the sampling parameters max\_new\_tokens = 512, top\_k = 50, top\_p = 0.8.

\paragraph{ICL + our auxiliary data} 
The prompt for in context learning  is shown in Table \ref{tab:prompts}. For generating answers to questions, this prompt is prepended to the appropriate evaluation prompt. 

\subsection{Auxiliary data generation}
For auxiliary task data generation we use the prompts shown in Table \ref{tab:prompts}. We set temperature to 0.75 and all other sampling parameters are set to the default in huggingface.

\begin{longtable}{p{\textwidth}}
    \caption{Prompt templates for evaluation and auxiliary data generation. 
    The sample model input provides a specific example input given to a model (chat tags are excluded). 
    In the evaluation prompt we replace question with image specific question. 
    In the auxiliary task data generation prompt we provide each of the task prompts listed in Figure \ref{fig:auxiliary_tasks}. 
    For in context learning, we concatenate the auxiliary task responses and provide it in context to the model. 
    In addition to all these prompts the image is passed to the model for generation.}
    \label{tab:prompts} \\

    \toprule
    \textbf{Prompt objective:} Evaluate a model on GQA, VQAV2, VQA-Rad \\
    \midrule
    \textbf{Prompt template:}  \{question\} Answer the question using a single word or phrase. \\ \\
    \textbf{Sample model input:} What is behind the white bench? Answer the question using a single word or phrase \\
    \midrule

    \textbf{Prompt objective:}  Evaluate a model on MMMU \\
    \midrule
    \textbf{Prompt template:} \{question\}

Answer the question above by strictly following the guidelines below. Your main goal is to provide the correct answer in the response. Do not deviate from the guidelines below.

1. Be Concise  
   - Provide a single word or brief phrase for the answer whenever possible, adhering to the final answer format.

2. Multiple-Choice (A, B, C, D)  
   - Respond only with the correct letter in square brackets, for example, [A].

3. If Reasoning is Needed  
   - Do not end your response with reasoning alone; always include the final answer as specified below.

4. Final Answer Format  
   - The correct answer must appear on the last line, preceded by the text: "Correct answer:"  
   - The answer should be either a single word/phrase or a letter in square brackets (e.g., [A]).  
   - If unsure, provide your best logical guess. \\

~\\
\textbf{Sample model input: }  
Margaret Gere's [image 1] was made in which medium?  
(A) Egg tempera  
(B) Watercolour  
(C) Ink  
(D) Oil paint  

Answer the question above by strictly following the guidelines below... (same rules as above). \\
\midrule

\textbf{Prompt objective:} Generate auxiliary task data \\
\midrule
\textbf{Prompt template:} \{Auxiliary task prompt\} \\ \\
\textbf{Sample model input:} What objects or people are visible in this image? \\
\midrule

\textbf{Prompt objective:} In context learning with our auxiliary data \\
\midrule
\textbf{Prompt template:} Here are a detailed list of captions of the image: \{concatenated\_captions\}. Answer the following question using these captions. \\ \\
\textbf{Sample model input:} Here are a detailed list of captions of the image: The image shows a suitcase, mask and table. This image has a table placed in front of suitcase... Answer the following question using these captions. \\
\bottomrule

\end{longtable}

\end{document}